\documentclass[journal]{IEEEtran}
\ifCLASSINFOpdf
  \usepackage[pdftex]{graphicx}
\else
  \usepackage[dvips]{graphicx}
\fi
\usepackage{amsmath}

\usepackage{color}

\usepackage[normalem]{ulem}

\begin{document}
%
\title{So Cloze yet so Far: N400 Amplitude is Better Predicted by Distributional Information than Human Predictability Judgements}
%
%
%

\author{James~A.~Michaelov,
    Seana~Coulson,
    and~Benjamin~K.~Bergen
\thanks{This work was partially supported by a 2020-2021 Center for Academic Research and Training in Anthropogeny Fellowship awarded to J.A. Michaelov.}
\thanks{J.A. Michaelov, S. Coulson, and B.K. Bergen are with the Department
of Cognitive Science, University of California San Diego, La Jolla, CA 92093 USA (email: j1michae@ucsd.edu).}%
\thanks{\textcopyright 2021 IEEE. Personal use of this material is permitted. Permission from IEEE must be obtained for all other uses, in any current or future media, including reprinting/republishing this material for advertising or promotional purposes, creating new collective works, for resale or redistribution to servers or lists, or reuse of any copyrighted component of this work in other works.}
\thanks{Published paper DOI: 10.1109/TCDS.2022.3176783}
}

%
%

\markboth{IEEE TRANSACTIONS ON COGNITIVE AND DEVELOPMENTAL SYSTEMS,~Vol.~X, No.~X, MONTH~YEAR}%
{Michaelov \MakeLowercase{\textit{et al.}}: So Cloze yet so Far}
%



\maketitle

\begin{abstract}
More predictable words are easier to process---they are read faster and elicit smaller neural signals associated with processing difficulty, most notably, the N400 component of the event-related brain potential. Thus, it has been argued that prediction of upcoming words is a key component of language comprehension, and that studying the amplitude of the N400 is a valuable way to investigate the predictions we make. In this study, we investigate whether the linguistic predictions of computational language models or humans better reflect the way in which natural language stimuli modulate the amplitude of the N400. One important difference in the linguistic predictions of humans versus computational language models is that while language models base their predictions exclusively on the preceding linguistic context, humans may rely on other factors. We find that the predictions of three top-of-the-line contemporary language models---GPT-3, RoBERTa, and ALBERT---match the N400 more closely than human predictions. This suggests that the predictive processes underlying the N400 may be more sensitive to the statistics of language than previously thought.
\end{abstract}


%
\IEEEpeerreviewmaketitle

\section{Introduction}
%
%
%
%
\IEEEPARstart{W}{hile} it is widely accepted that predictable words are easier to process than unpredictable ones, the role of predictive processes in language comprehension has long been an issue of contentious debate (for reviews, see \cite{kutas_2011_LookWhatLies,vanpetten_2012_PredictionLanguageComprehension,luke_2016_LimitsLexicalPrediction,kuperberg_2016_WhatWeMean}). One prominent position is that the language processor does not waste resources on predictive processing \cite{forster_1981_PrimingEffectsSentence}. Under such an account, because there are an infinite number of possible continuations for any given natural language string, linguistic predictions would be wrong far more often than they would be right. Thus, given the limited value of linguistic prediction, the language processor simply does not engage in it \cite{jackendoff_2002_FoundationsLanguageBrain}. Advocates of this position have attributed observed predictability effects on language processing to the demands of integrating the meaning of a word into its preceding context \cite{schwanenflugel_1985_InfluenceSentenceConstraint,traxler_2000_EffectsSentenceConstraint}, some form of automatic spreading activation in the lexicon \cite{west_1982_SourceInhibitionExperiments,collins_1975_SpreadingactivationTheorySemantic}, or both. 

However, there is growing evidence in support of prediction as a component of language comprehension. Much of this research comes from looking at neural signals of processing difficulty, especially the N400, a negative-going component of the event-related brain potential (ERP) that peaks roughly 400ms after the presentation of a meaningful stimulus \cite{kutas_1980_ReadingSenselessSentences,kutas_2011_ThirtyYearsCounting}. With linguistic stimuli, the size of the N400 is sensitive to semantic congruity---N400 amplitude is large by default, and is reduced if the word is facilitated by the preceding context \cite{vanpetten_2012_PredictionLanguageComprehension,delong_2020_ComprehendingSurprisingSentences,kuperberg_2020_TaleTwoPositivities}. In recent years, a range of studies have found that N400 amplitude modulations appear to reflect lexical properties of specific nouns that are semantically predictable; thus, researchers have argued that N400 predictability effects do not simply reflect ease of integration or spreading activation, and---at least some of the time---provide evidence for predictive processes in language comprehension 
\cite{delong_2005_ProbabilisticWordPreactivation,vanberkum_2005_AnticipatingUpcomingWords,otten_2007_GreatExpectationsSpecific,kwon_2017_PredictingSemanticFeatures,kuperberg_2020_TaleTwoPositivities,nicenboim_2020_AreWordsPreactivated,urbach_2020_ExploratoryDataAnalysis,fleur_2020_DefinitelySawIt}.

What are these predictions based on? Since the early days of N400 research, cloze probability \cite{taylor_1953_ClozeProcedureNew} has served as the chief metric of contextual word predictability \cite{kutas_1984_BrainPotentialsReading,vanpetten_2012_PredictionLanguageComprehension,brothers_2021_WordPredictabilityEffects}. The cloze probability of a given word is defined as the proportion of people who fill a gap in a sentence with that specific word \cite{taylor_1953_ClozeProcedureNew}, and thus, provides a measure of how predictable a word is in a specific sentence context. It is well-established that words with a higher cloze probability elicit a smaller N400 response compared to words with lower cloze probabilities \cite{kutas_1984_BrainPotentialsReading,kutas_2011_ThirtyYearsCounting,kuperberg_2020_TaleTwoPositivities}, as well as being read faster and recognized faster \cite{brothers_2021_WordPredictabilityEffects}---in fact, some work has shown that cloze probability and N400 amplitude are inversely correlated at a level of over 90\% \cite{kutas_1994_PsycholinguisticsElectrifiedEventrelated}. A more recent operationalization of predictability is derived from language models (LMs), computational systems designed to predict a word in context. Unlike humans, these LMs are only trained on text data as input, and consequently base their predictions solely on the statistics of language \cite{jurafsky_2021_SpeechLanguageProcessing}. Thus, while linguistic predictions in humans may utilize a range of knowledge both linguistic and extra-linguistic, LMs learn the actual distributional probability of a word in context in the corpus on which they are trained \cite{smith_2011_ClozeNoCigar,brothers_2021_WordPredictabilityEffects}.

Understanding the relationship between N400 amplitude and the statistics of language is vital to understanding the N400 \cite{michaelov_2020_HowWellDoes}. Given the evidence that N400 amplitude is affected by linguistic input over the lifespan \cite{kutas_2011_ThirtyYearsCounting}, and the fact that they are models trained purely on linguistic input, LMs give us a precise way to model the extent to which linguistic input alone can predict the N400 response. On the other hand, there is no way to tell which sources of information and neurocognitive processes are involved when experimental participants complete the cloze task. Thus, even if cloze probability were to correlate more closely with N400 amplitude than LM predictions, it is less informative in terms of illuminating the basis of prediction in language comprehension.

However, recent work suggests that this trade-off between accuracy and explainability may be nearing an end. The statistics of language---as operationalized by LM predictions---can not only successfully predict single-trial N400 amplitudes \cite{frank_2015_ERPResponseAmount,aurnhammer_2019_EvaluatingInformationtheoreticMeasures,merkx_2021_HumanSentenceProcessing,michaelov_2021_DifferentKindsCognitive} and the significant differences in N400 amplitude elicited by a range of experimental manipulations \cite{michaelov_2020_HowWellDoes}, but at least for some stimuli may be better at this than cloze probability \cite{michaelov_2020_HowWellDoes,michaelov_2021_DifferentKindsCognitive}. However, the two studies in which LM predictions outperform cloze have either looked at the effects without direct comparison to the N400 data \cite{michaelov_2020_HowWellDoes} or targeted data from an experiment intended to show the N400 responds to factors other than cloze \cite{michaelov_2021_DifferentKindsCognitive}. 

The goal of the present study is to test whether the amplitude of the N400 to words in sentence contexts can be better predicted by the statistics of language than by cloze probability---even under conditions that are maximally favorable to cloze. Using ERP data from a large-scale multiple-laboratory experiment \cite{nieuwland_2018_LargescaleReplicationStudy}, we used linear mixed effects regression models to examine how well the amplitude of the N400 elicited by experimental stimuli was predicted by the cloze probabilities gathered in the original experiment \cite{nieuwland_2018_LargescaleReplicationStudy}, and compared its performance to that of several pretrained neural network LMs \cite{gulordava_2018_ColorlessGreenRecurrent,jozefowicz_2016_ExploringLimitsLanguage,devlin_2019_BERTPretrainingDeep,liu_2019_RoBERTaRobustlyOptimized,radford_2019_LanguageModelsAre,dai_2019_TransformerXLAttentiveLanguage,lan_2020_ALBERTLiteBERT,brown_2020_LanguageModelsAre}. Language models are the best way to capture prediction based on language statistics at present. If any contemporary models predict N400 amplitude better than cloze probability does, that would constitute compelling evidence that prediction, as measured by the N400, can be driven by language statistics.

\section{Background}

\subsection{Cloze probability}
Cloze probability has long been used to asses a word's predictability in context \cite{vanpetten_2012_PredictionLanguageComprehension,delong_2014_PreProcessingSentenceComprehension,luke_2016_LimitsLexicalPrediction,brothers_2021_WordPredictabilityEffects}. In addition to its use in understanding the N400 \cite{kutas_1984_BrainPotentialsReading,kutas_2011_ThirtyYearsCounting}, it has been shown to predict behavioural correlates of processing difficulty, such as word reading time \cite{brothers_2021_WordPredictabilityEffects}. In fact, when directly compared, cloze probability has previously be found to be better at predicting such behavioural metrics than LMs \cite{brothers_2021_WordPredictabilityEffects}.

However, while cloze probability is a metric grounded in human judgements, it may not be as helpful in understanding online human comprehension as might appear at first glance. As discussed, predictability effects are thought to arise from individuals' graded predictions about upcoming words, whereas cloze probability is an aggregate measure over a sample of individuals based exclusively on their top prediction. In addition to the question of whether we should expect these two distributions to be equivalent, there is also a practical issue of sample size---less likely continuations require a larger sample of individuals in order for even a single experimental participant to produce. Indeed, as a language production task, its relevance for comprehension is unclear in view of disagreement regarding the extent of overlap between the production and comprehension systems (see \cite{meyer_2016_SameDifferentClosely,hendriks_2014_AsymmetriesLanguageProduction} for review and discussion), it is not necessarily the case that the next-word probability of a word will be the same for both the production and comprehension system. 

Beyond these concerns, and even if cloze is a good predictor of processing difficulty due to predictability overall (e.g. as measured by reading time), when investigating the N400, the temporal dimension must also be considered. Cloze probability is based on responses produced by experimental participants after reading a sentence with a gap that must be filled in. Given the substantial evidence that there are neurocognitive processes involved in human language comprehension that occur after the N400 \cite{delong_2020_ComprehendingSurprisingSentences,kuperberg_2020_TaleTwoPositivities}, even if it is the case that the N400 and cloze probability both reflect individuals' graded predictions, and that cloze responses are influenced by the predictions that underlie the N400 response, it should not be taken as a given that these predictions are the same. Thus, there is no \textit{a priori} reason to assume that cloze probability is the best possible operationalization of the predictions that underlie the N400. 

\subsection{Language model predictions}
LMs are trained to predict the probability of a word based only on the linguistic context. Given that such models do not explicitly learn meanings of words, and that the N400 response to a word is thought to be largely or wholly determined by meaning \cite{kutas_2011_ThirtyYearsCounting,kuperberg_2020_TaleTwoPositivities}, intuitively, we may expect them to perform poorly at predicting the amplitudes of N400 responses to words. However, previous research has shown that LMs can learn semantic relationships between words \cite{rogers_2021_PrimerBERTologyWhat}. Thus, the extent to which LMs can acquire semantic knowledge, and specifically, knowledge about the semantic relations between words, may be greater than would be expected \textit{prima facie}. Whether or not humans can learn quite so much based on only linguistic input is an open question, but there is evidence that we may learn semantic relations between referents of words with which we have no direct experience \cite{marmor_1978_AgeOnsetBlindness}.

An additional benefit of using LM predictions to operationalize word predictability is that researchers know exactly what sources of information are used by these models---they are trained on specific data, and thus researchers can form hypotheses about how the specific kinds of information in these data may be used to predict upcoming linguistic input, and by which system. This is especially important given that, as discussed, we might expect the predictions underlying the N400 to also impact cloze probability. If factors beyond linguistic input such as world knowledge have an effect on N400 amplitude, as has been proposed \cite{kutas_2011_ThirtyYearsCounting}, then they are also likely to have an effect on cloze probability. For this reason, when using cloze probability to predict N400 amplitude, it may be impossible to disentangle the effect of each source of information, and thus limiting the extent to which we can understand the basis upon which the predictions underlying the N400 are made. Using metrics based on the statistics of language (for example, LM predictions) may therefore be one of the only ways to successfully isolate the specific effect of linguistic input on N400 amplitude.

\subsection{Language model surprisal}
When LM predictions are used to investigate predictability effects on language comprehension, predictability is usually not operationalized as the raw probability of words as calculated by these models, but rather, their surprisal. The surprisal $S$ of a word $w_i$ is the negative logarithm of its probability given its preceding context $w_1...w_{i-1}$, as shown in (\ref{eq:surprisal}).
\begin{equation}
  S(w_{i}) = -\log P(w_{i}|w_{1}...w_{i-1})
  \label{eq:surprisal}
\end{equation}
In addition to theoretical claims behind surprisal theory as an explanation of predictability effects in language comprehension \cite{hale_2001_ProbabilisticEarleyParser,levy_2008_ExpectationbasedSyntacticComprehension,smith_2013_EffectWordPredictability}, there is also an array of evidence showing that LM surprisal correlates with behavioural metrics of processing difficulty such as reading time
\cite{boston_2008_ParsingCostsPredictors,demberg_2008_DataEyetrackingCorpora,roark_2009_DerivingLexicalSyntactic,mitchell_2010_SyntacticSemanticFactors,smith_2011_ClozeNoCigar,monsalve_2012_LexicalSurprisalGeneral,willems_2016_PredictionNaturalLanguage}. 
A further body of research has found that LM surprisal is a significant predictor of N400 amplitude, with the surprisal of generally better-performing and more advanced LMs showing a better fit to the N400 data \cite{frank_2015_ERPResponseAmount,aurnhammer_2019_EvaluatingInformationtheoreticMeasures,merkx_2021_HumanSentenceProcessing,michaelov_2021_DifferentKindsCognitive}. Additionally, when LMs are given the same experimental stimuli as humans in neurolinguistic experiments, significant differences in surprisal often match significant differences in N400 as a function of experimental condition---again, with generally better-performing and more advanced models matching the human responses better \cite{michaelov_2020_HowWellDoes,michaelov_2021_DifferentKindsCognitive}.

In previous work, operationalizing predictability as cloze probability generally appears to yield better results for human behavioural data than LM surprisal \cite{brothers_2021_WordPredictabilityEffects}; however, this has not been well-explored for the N400. To the best of our knowledge, only one published paper has directly compared how well cloze probability and LM surprisal predict N400 amplitude, finding that LM surprisal performs better \cite{michaelov_2021_DifferentKindsCognitive}. However, the comparison between cloze probability and LM prediction was not an aim of that previous study, and thus there are several caveats to be noted about this result. Firstly, the study investigated the N400 response to words with the same cloze probability but which were either related or unrelated to the highest-cloze completion---there is a well-established effect showing that the former elicit lower-amplitude N400s than the latter \cite{kutas_1984_BrainPotentialsReading,kutas_1993_CompanyOtherWords,federmeier_1999_RoseAnyOther,thornhill_2012_LexicalConceptualAnticipation,ito_2016_PredictingFormMeaning}. Thus, cloze is inherently at a disadvantage in prediction, given that the two conditions are controlled for cloze. The study also involved a condition where all stimuli had a cloze of zero; thus, none of the variance in N400 amplitude within this condition could be explained by cloze. Finally, the study compared raw cloze probability to LM surprisal---given that the surprisal calculated from cloze probability has been found to correlate with behavioural predictability effects \cite{smith_2011_ClozeNoCigar,lowder_2018_LexicalPredictabilityNatural}, a fair comparison would also involve cloze surprisal. The finding that surprisal can differ between words that are matched for cloze but either related or unrelated to the highest-cloze continuation of a sentence is also found in another study \cite{michaelov_2020_HowWellDoes}, but this study only compares significant differences in surprisal to the significant differences reported in the original papers---there is no direct comparison made between the surprisal and N400 data.

\subsection{The present study}
In the present study, we aim to provide just such a fair comparison using modern LMs and openly available data from a large N400 study (n = 334) \cite{nieuwland_2018_LargescaleReplicationStudy}. First, we use data from a study that was specifically designed to investigate the effect of cloze probability on N400 amplitude; thus, there are none of the aforementioned cases where experimental conditions are matched by cloze and differ in another way (that may be reflected in LM predictions, see \cite{michaelov_2020_HowWellDoes,michaelov_2021_DifferentKindsCognitive}). Additionally, we remove the data from all stimuli with a cloze probability of zero. Given that previous work has shown that there is variability in N400 amplitude between experimental conditions where all items had a cloze probability of zero \cite{metusalem_2012_GeneralizedEventKnowledge,ito_2016_PredictingFormMeaning}, and some of these studies have been successfully modeled using LM predictions \cite{michaelov_2020_HowWellDoes}, there is a chance that including these would give the LMs an unfair advantage. Finally, we compare both raw cloze probability and cloze surprisal to ensure that the log-transformation of LM probability is not a confound, as previous work has suggested that there may be a logarithmic linking function between human-derived metrics of word probability and processing difficulty \cite{smith_2011_ClozeNoCigar,lowder_2018_LexicalPredictabilityNatural,delaney-busch_2019_NeuralEvidenceBayesian}.

\section{Method}
\subsection{Original study and data}
We use EEG data from a large-scale experiment by Nieuwland and colleagues \cite{nieuwland_2018_LargescaleReplicationStudy}. In this experiment, participants read sentences one word at a time, with ERPs time-locked to previously-determined target words. In the data provided, the N400 is operationalized as the mean amplitude voltage recorded from the centro-parietal region of interest (electrodes Cz, C3, C4, Pz, P3, and P4) 200--500ms after the presentation of the target word. We use the data provided for target nouns, which replicate the well-established finding that higher-cloze nouns elicit smaller (less negative) N400 responses than lower-cloze nouns \cite{nieuwland_2018_LargescaleReplicationStudy,kutas_1984_BrainPotentialsReading,kutas_2011_ThirtyYearsCounting}. 

To calculate the cloze probability of items in the original study, each stimulus sentence was truncated before the target word \cite{nieuwland_2018_LargescaleReplicationStudy}. Thus, participants in the cloze task were presented with the preceding linguistic context for the target word and asked to complete the sentence. The cloze probabilities were then calculated on the basis of the responses from two sets of 30 participants, each of which completed the cloze task for half of the total stimulus sentences. The authors provide both the cloze and ERP data online (at https://osf.io/eyzaq/).

The electrophysiological experiment was carried out at 9 laboratories in the United Kingdom and comprises data from 334 participants, reaching a total of 25,849 trials. We removed all items with a cloze probability of zero for fair comparison with LM surprisal, as previously discussed. Finally, we used the cloze data to calculate cloze surprisal for each remaining item. Because all zero-cloze items were removed, this also removed the need for smoothing zero-probabilities, as has been done in previous related work \cite{lowder_2018_LexicalPredictabilityNatural}.

\subsection{Language models}
We operationalize corpus-based probability of a word in context as the probability calculated by a neural network LM. There are many different architectures for neural network LMs, some of which have been used to model behavioural and neural correlates of human language processing. Here we focus on the two most prolific and successful types of LM in recent years---RNNs and transformers.

\subsubsection{RNNs}

Until the development of transformer LMs \cite{vaswani_2017_AttentionAllYou}, recurrent neural network (RNN) language models long dominated the field. With their memory bottleneck and their incremental processing of words \cite{keller_2010_CognitivelyPlausibleModels,merkx_2021_HumanSentenceProcessing}, RNNs have often been used as cognitive models of human language processing \cite{elman_1990_FindingStructureTime}, including prior efforts to model the N400 \cite{frank_2015_ERPResponseAmount,aurnhammer_2019_EvaluatingInformationtheoreticMeasures,michaelov_2020_HowWellDoes,merkx_2021_HumanSentenceProcessing,michaelov_2021_DifferentKindsCognitive}. In the present study, we use two RNN LMs referred to in the literature (see, e.g., \cite{futrell_2019_NeuralLanguageModels}) as GRNN \cite{gulordava_2018_ColorlessGreenRecurrent} and JRNN \cite{jozefowicz_2016_ExploringLimitsLanguage}. Previous research has found JRNN surprisal to more closely resemble N400 amplitude than does GRNN surprisal \cite{michaelov_2020_HowWellDoes}. GRNN and JRNN surprisal were calculated using the code accompanying Michaelov and Bergen \cite{michaelov_2020_HowWellDoes}.

\subsubsection{Transformers}

Transformer language models are a neural network LM architecture \cite{vaswani_2017_AttentionAllYou} that has been found to outperform RNNs at the standard language modeling task (predicting words from context, see \cite{dai_2019_TransformerXLAttentiveLanguage} for review), as well as a range of other tasks \cite{devlin_2019_BERTPretrainingDeep,radford_2019_LanguageModelsAre}. Transformer LMs have also been shown to do better than RNNs at predicting N400 amplitude \cite{merkx_2021_HumanSentenceProcessing,michaelov_2021_DifferentKindsCognitive}. The present study includes two varieties of transformer LMs---\textit{autoregressive language models} trained on the traditional task of predicting words based on their preceding linguistic context, and \textit{masked language models}, trained to fill a gap in a sentence, and that thus can use words that appear both before and after in its prediction of the target word. We include the probabilities from three autoregressive LMs in our analysis---Transformer-XL \cite{dai_2019_TransformerXLAttentiveLanguage}, GPT-2 \cite{radford_2019_LanguageModelsAre}, and GPT-3 \cite{brown_2020_LanguageModelsAre}. The three masked LMs that we use to calculate word probability are BERT \cite{devlin_2019_BERTPretrainingDeep}, RoBERTa \cite{liu_2019_RoBERTaRobustlyOptimized}, and ALBERT \cite{lan_2020_ALBERTLiteBERT}. For all transformer LMs except for GPT-3, we use the implementation of each model made available through the \textit{transformers} \cite{wolf_2020_TransformersStateoftheArtNatural} package to calculate surprisal. GPT-3 predictions were accessed via the OpenAI API \cite{openai_2021_OpenAIAPI}.

\begin{table}[h]
\renewcommand{\arraystretch}{1.2}
\caption{Summary of language models used}
\label{tab:models}

{\centering

\begin{tabular}{lrrr}
\hline
\textbf{Model}   & \textbf{Parameters}\footnote{} & \textbf{Corpus size}\footnote{} & \textbf{Ref.} \\ \hline
GRNN        & 71.8M        & 90M         & \cite{gulordava_2018_ColorlessGreenRecurrent} \\
JRNN        & 1.04B        & 1B         & \cite{jozefowicz_2016_ExploringLimitsLanguage} \\
Transformer-XL\footnote{}    & 285M        & 103M        & \cite{dai_2019_TransformerXLAttentiveLanguage} \\
GPT-2 (XL)     & 1.56B        & $\sim$8B      & \cite{radford_2019_LanguageModelsAre} \\
GPT-3 (Davinci)   & 175B        & $\sim$300B      & \cite{brown_2020_LanguageModelsAre} \\
BERT (large, cased, WWM\footnote{}) & 334M        & 3.3B        & \cite{devlin_2019_BERTPretrainingDeep} \\
RoBERTa (large)   & 355M        & $\sim$33B      & \cite{liu_2019_RoBERTaRobustlyOptimized} \\
ALBERT (XXLarge v2)\footnote{} & 206M        & 3.3B        & \cite{lan_2020_ALBERTLiteBERT} \\ \hline
\end{tabular}

}

\vspace{0.5em}

\footnotesize{$^1$ The number of free parameters for the \textit{transformers} \cite{wolf_2020_TransformersStateoftheArtNatural} implementations of Transformer-XL, GPT-2, BERT, RoBERTa, and ALBERT were calculated using \textit{pytorch} \cite{paszke_2019_PyTorchImperativeStyle}. For JRNN and GPT-3, we utilized the models directly provided by the authors of the paper, and so use the number of parameters reported in the cited paper or its supplementary materials \cite{jozefowicz_2016_ExploringLimitsLanguage,brown_2020_LanguageModelsAre}. While we use the author-provided GRNN, no estimate of model parameters is given in the original paper \cite{gulordava_2018_ColorlessGreenRecurrent}, so we calculated this with \textit{pytorch} \cite{paszke_2019_PyTorchImperativeStyle}. 

$^2$ Number of words in training corpus is reported in the original papers \cite{gulordava_2018_ColorlessGreenRecurrent,jozefowicz_2016_ExploringLimitsLanguage,dai_2019_TransformerXLAttentiveLanguage,devlin_2019_BERTPretrainingDeep}, or estimated (denoted by `$\sim$'). ALBERT is trained on the same data as BERT \cite{lan_2020_ALBERTLiteBERT}. Training data for GPT-2 and RoBERTa are estimated based on a comparison of file size with the dataset used for BERT. GPT-3 is trained on 300 billion tokens; however, given that it uses byte-pair encoding for tokenization \cite{brown_2020_LanguageModelsAre,radford_2019_LanguageModelsAre,sennrich_2016_NeuralMachineTranslation}, the actual number of words is lower.

$^3$ We use the \textit{transformers} \cite{wolf_2020_TransformersStateoftheArtNatural} implementation of Transformer-XL; some models reported in the original paper \cite{dai_2019_TransformerXLAttentiveLanguage} have a higher number of parameters.

$^4$ Whole-word masking, see \cite{googleresearch_2018_BERT}.

$^5$} Note that while ALBERT has fewer free parameters than either BERT or RoBERTa, it shares parameters between layers, and so is actually a much larger model than either BERT or RoBERTa \cite{lan_2020_ALBERTLiteBERT}.
\end{table}

\subsection{Language model predictions}
The aforementioned LMs were thus used to predict the probability of the target nouns from the original study \cite{nieuwland_2018_LargescaleReplicationStudy}. Each stimulus sentence was truncated before the target word and the predicted probabilities generated by the models for each of the target words were recorded. Thus, all the models, including the masked LMs, were required to base their predictions on the preceding context. This procedure was intended to match the cloze task, where sentences were truncated in the same way, as well as the ERP experiment, where experimental participants had read only the preceding context when they reached the target word. These probabilities were then transformed into surprisals using the formula in (\ref{eq:surprisal}). We used a logarithm of base 2 so that surprisal can be measured in bits \cite{futrell_2019_NeuralLanguageModels}. For fair comparison, only words appearing in all models' vocabularies were included in the analysis.

\subsection{Predicting the N400}
The LM surprisal values, original cloze values, cloze surprisal values, and by-trial N400 amplitudes were all z-transformed before running statistical analyses. These z-transformed LM surprisals, cloze surprisals, and cloze probabilities were then used to predict the z-transformed by-trial N400 amplitudes. After the removal of data for all target words that either did not appear in all LMs' vocabularies or that had a cloze probability of zero, our final dataset consisted of N400 data from 15,551 trials, elicited by 94 different sentences.

Statistical analysis and data manipulation were carried out in \textit{R} \cite{rcoreteam_2020_LanguageEnvironmentStatistical} using \textit{Rstudio} \cite{rstudioteam_2020_RStudioIntegratedDevelopment} and the \textit{tidyverse} \cite{wickham_2019_WelcomeTidyverse}, \textit{lme4} \cite{bates_2015_FittingLinearMixedeffects}, and \textit{ggh4x} \cite{vandenbrand_2021_Ggh4xHacksGgplot2} packages, and the code provided by Nicenboim et al. \cite{nicenboim_2020_AreWordsPreactivated} for preparing the data \cite{nieuwland_2018_LargescaleReplicationStudy}. To reduce the risk of Type I errors, all $p$-values in our analyses are corrected for multiple comparisons based on false discovery rate \cite{benjamini_2001_ControlFalseDiscovery}.

\section{Results}

\subsection{Preliminary analysis with cloze probability}
First, we test whether the original finding, that higher-cloze nouns elicit smaller N400s than lower-cloze nouns, still holds for our subset of the data. We did this by following the original statistical methods as closely as possible \cite{nieuwland_2018_LargescaleReplicationStudy}. For this reason, we used linear mixed-effects regression models with the same covariates as in the original analyses; and in order to test the significance of variables, we use likelihood ratio tests on nested regressions.

After running all regressions (including those described in the following subsections), we found that including the original random effect structure of random slopes for experimental participant and item resulted in singular fits in several cases; so these were reduced to random intercepts in all models. Following the original analysis, we also included the laboratory in which the experiment was run as a fixed effect. 

As in the original study, we found no interaction between cloze probability and laboratory ($\chi^2(8) = 7.357, p = 1$). However, unlike the original study, we found a significant effect of laboratory even when controlling for cloze probability ($\chi^2(8) = 36.280, p < 0.001$). This may be due to the difference in sample or in random effects structure. Crucially, we found a significant effect of cloze probability even when controlling for laboratory ($\chi^2(1) = 27.937, p < 0.001$). Thus, we replicated the noun predictability effect on our subset of the data.

\subsection{Cloze surprisal and N400 amplitude}
Running the same tests with cloze surprisal (i.e. negative log-transformed cloze probability) replacing cloze probability leads to the same results (Cloze surprisal x lab: $\chi^2(8) = 3.596, p = 1$; cloze surprisal: $\chi^2(1) = 29.403, p < 0.001$; lab: $\chi^2(8) =  36.241, p < 0.001$). Thus, we included laboratory as a covariate for our remaining analyses.

To compare cloze probability and cloze surprisal as predictors of N400, we compared the two best regressions including each as a main effect---namely, those also including laboratory as a main effect but not the interaction between the two. Since the two regressions are not nested, we employed Akaike's Information Criterion (AIC) \cite{akaike_1973_InformationTheoryExtension} to compare them. We found that the regression with cloze surprisal as a fixed effect has a slightly lower AIC (AIC = 113227.2) than the regression with cloze probability as a fixed effect (AIC = 113228.7). 

These AIC values can be used to calculate evidence ratios based on Akaike weights (see \cite{wagenmakers_2004_AICModelSelection}). Based on this approach, we find that with an evidence ratio of 2.08, the cloze surprisal regression is 2.08 times more likely than the cloze probability regression to be the best model of the N400 data. 

However, when comparing AIC values, a general rule of thumb is that when there is an absolute difference in AIC of 2 or less between two statistical models, they have similar levels of support, while a difference of 4 or more means that the model with a lower AIC has `considerably' more evidential support \cite{burnham_2004_MultimodelInferenceUnderstanding}. In this case, the cloze surprisal regression has an AIC which is 1.47 less than the cloze probability regression. Thus, despite the evidence ratio of 2.08, the two regressions should be considered to have similar levels of support, and so it is still not clear whether cloze probability or cloze surprisal is a better predictor of N400 amplitude.

In order to investigate this further, we ran additional analyses, finding that that the two explain the same variance in N400 amplitude: adding cloze surprisal to the best cloze probability regression does not improve model fit ($\chi^2(1) = 1.638, p = 0.965$); and neither does adding probability to the best cloze surprisal regression ($\chi^2(1) = 0.171, p = 1$). However, given the lower (i.e., better) AIC of the cloze surprisal regression, we take cloze surprisal as the most explanatory representation of cloze for the remainder of our analyses.

\subsection{Language model surprisal and N400 amplitude}
We calculated the probability of each target word based on the predictions of GRNN (mean = $0.087$; standard deviation = $0.190$), JRNN ($0.211 \pm 0.291$), Transformer-XL ($0.092 \pm 0.192$), GPT-2 ($0.382 \pm 0.358$), GPT-3 ($0.526 \pm 0.371$), BERT ($0.317 \pm 0.355$), RoBERTa ($0.495 \pm 0.374$), and ALBERT ($0.298 \pm 0.316$) for comparison with cloze ($0.631 \pm 0.348$). These probabilities were then transformed into surprisal.

We tested whether the surprisal calculated from each LM is a significant predictor of N400 amplitude. To do this, we compared regressions with a main effect of laboratory and random intercepts for subject and item to those also including a main effect of the relevant LM's surprisal. In this way, the analysis matches those investigating the main effect of cloze probability and cloze surprisal. The results of these analyses are shown in Table \ref{tab:null_surprisal}. As can be seen, main effects of surprisal calculated using JRNN, Transformer-XL, GPT-2, GPT-3, BERT, RoBERTa, and ALBERT are all significant in their respective regressions, but the main effect of GRNN surprisal is only marginally significant.

\begin{table}[h]
\renewcommand{\arraystretch}{1.2}
\caption{Significant predictors of N400 amplitude}
\label{tab:null_surprisal}
\centering
\begin{tabular}{lrrr}
\hline
\textbf{Predictor}      & $\boldsymbol{\chi^2(\textbf{df}=1)}$ & \textbf{p}        \\ \hline
GRNN surprisal          & 6.356          & 0.072           \\
\textbf{JRNN surprisal}     & \textbf{17.330}  & \textbf{\textless{}0.001}      \\
\textbf{Tranformer-XL surprisal} & \textbf{19.158}  & \textbf{\textless{}0.001} \\
\textbf{GPT-2 surprisal}     & \textbf{26.313}  & \textbf{\textless{}0.001} \\
\textbf{GPT-3 surprisal}     & \textbf{40.817}  & \textbf{\textless{}0.001} \\
\textbf{BERT surprisal }         & \textbf{30.760}           & \textbf{\textless{}0.001}            \\
\textbf{RoBERTa surprisal}    & \textbf{37.848}  & \textbf{\textless{}0.001} \\
\textbf{ALBERT surprisal}    & \textbf{35.918}  & \textbf{\textless{}0.001} \\ \hline
\end{tabular}
\end{table}

\subsection{Comparison of model fit}
\begin{figure*}[!t]
\centering
\includegraphics[width=0.9\textwidth]{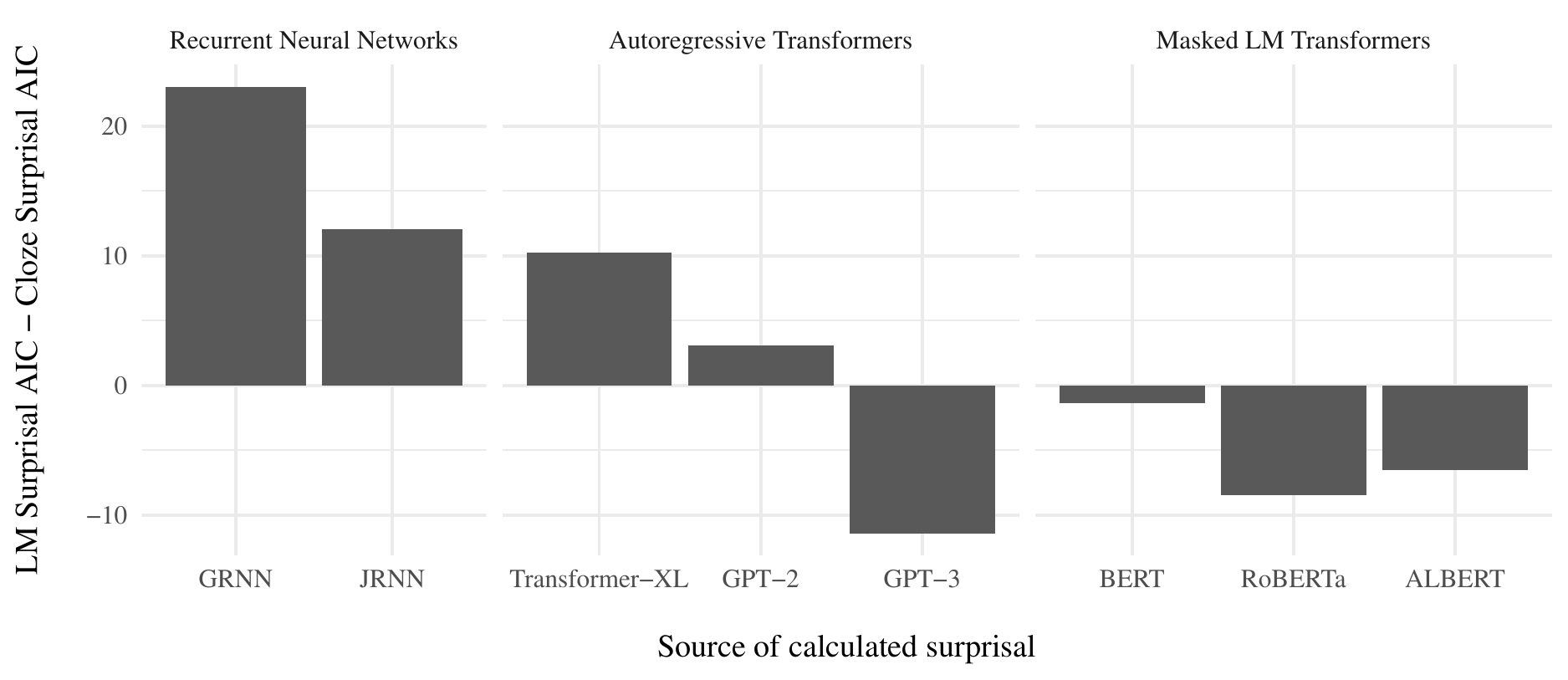}
\caption{AICs of all regressions including fixed effects of the denoted surprisal and laboratory, as well as random intercepts for each item and experimental participants. For easier comparison, AIC is scaled by subtracting the AIC of the regression including cloze surprisal, laboratory, and the aforementioned random intercepts. Lower AICs indicate better model fit \cite{akaike_1973_InformationTheoryExtension}.}
\label{fig:AICs}
\end{figure*}

We next compared the AICs of each linear mixed-effects regression model including LM surprisal with one that instead used cloze surprisal. These comparisons are presented in Figure \ref{fig:AICs}, which shows the AIC of each LM surprisal regression with the AIC of the cloze surprisal regression subtracted. This allows for easier comparison of regression AIC, and has a clear interpretation---any regression with a relative AIC of less than zero has a better fit than the cloze surprisal regression.

As can be seen in Figure \ref{fig:AICs}, the regressions based on the surprisals calculated from four LMs have lower AICs than cloze surprisal (AIC = 113227.2): GPT-3 (AIC = 113215.8; evidence ratio with cloze surprisal = 300.89), BERT (AIC = 113225.9; evidence ratio = 1.97), RoBERTa (AIC = 113218.8; evidence ratio = 68.18), and ALBERT (AIC = 3113220.7 ; evidence ratio = 25.98). The AIC of the remaining models is higher than that of cloze surprisal. It should be noted that in all but one case, the difference in AIC between the cloze surprisal and all other regressions is greater than 4, suggesting a meaningful difference in this respect \cite{burnham_2004_MultimodelInferenceUnderstanding}. The one exception is the BERT regression ($\Delta$AIC = 1.36)---thus, while the BERT regression is 1.97 times more likely than the cloze surprisal regression to provide the best fit to the N400 data, we rely on the tests in the rest of this section to determine whether BERT surprisal is in fact a better predictor of N400 amplitude than cloze surprisal.

In sum, regressions based on the surprisals derived from GPT-3, RoBERTa, and ALBERT more closely fit the N400 data than the regression based on cloze surprisal, and this may also be the case for the BERT surprisal regression.

\subsection{Does language model surprisal improve fit of regressions based on human cloze data?}
In addition to comparing the AICs of the models, following Brothers and Kuperberg \cite{brothers_2021_WordPredictabilityEffects}, we compared how well cloze and LM surprisal predict N400 amplitude by constructing additional regressions with both variables and comparing them to regressions with only one. First, we compared the effect of adding the surprisal calculated from each LM to a regression already including cloze surprisal. Thus, we tested whether each LM surprisal explains variance in N400 amplitude above and beyond that which is already explained by cloze surprisal. The results are shown in Table \ref{tab:cloze_vs_lms}.

\begin{table}[b]
\renewcommand{\arraystretch}{1.2}
\caption{Results of LRTs testing whether adding LM surprisal as a main effect improves the fit of regressions that already include cloze surprisal as main effect}
\label{tab:cloze_vs_lms}
\centering
\begin{tabular}{lrrr}
\hline
\textbf{Predictor} & $\boldsymbol{\chi^2(\textbf{df}=1)}$     & \textbf{p}   \\ \hline
GRNN surprisal        & 0.056           & 1       \\
JRNN surprisal        & 3.982           &  	0.260     \\
Tranformer-XL surprisal   & 3.031           & 0.424     \\
GPT-2 surprisal       & 5.088           & 0.142     \\
\textbf{GPT-3 surprisal}   & \textbf{12.168}  & \textbf{0.004} \\
\textbf{BERT surprisal}        & \textbf{9.639}           & \textbf{0.015}       \\
\textbf{RoBERTa surprisal}  & \textbf{11.720}  & \textbf{0.005} \\
\textbf{ALBERT surprisal}  & \textbf{8.450}  & \textbf{ 	0.026} \\ \hline
\end{tabular}
\end{table}

As can be seen in Table \ref{tab:cloze_vs_lms}, adding GPT-3, BERT, RoBERTa, or ALBERT surprisal to regressions already including cloze surprisal significantly improves their fit, while adding the surprisal of other LMs does not. 

\subsection{Does human cloze data improve fit of regressions based on language model surprisal?}
We also ran the reverse analysis, investigating the effect of adding cloze surprisal to a regression that already includes one LM surprisal as a fixed effect. Thus, we test whether cloze surprisal explains variance in N400 amplitude not explained by each LM surprisal. The results are shown in Table \ref{tab:lms_vs_cloze}.

\begin{table}[b]
\renewcommand{\arraystretch}{1.2}
\caption{Results of LRTs testing whether adding cloze surprisal as a main effect improves the fit of regressions that already include LM surprisal as main effect}
\label{tab:lms_vs_cloze}
\centering
\begin{tabular}{lrrr}
\hline
\textbf{Predictor} & $\boldsymbol{\chi^2(\textbf{df}=1)}$ & \textbf{p}    \\ \hline
\textbf{GRNN surprisal}        & \textbf{23.103}           & \textbf{\textless{}0.001} \\
\textbf{JRNN surprisal}        & \textbf{16.056}          & \textbf{0.001} \\
\textbf{Tranformer-XL surprisal}   & \textbf{13.277}          & \textbf{0.002}      \\
\textbf{GPT-2 surprisal }       & \textbf{8.178}          & \textbf{0.028}      \\
GPT-3 surprisal       & 0.754           & 1  \\
\textbf{BERT surprisal }       & \textbf{ 	8.282}          & \textbf{0.027} \\
RoBERTa surprisal      & 3.276           & 0.380  \\
ALBERT surprisal       & 1.935           & 0.820  \\ \hline
\end{tabular}
\end{table}

As can be seen in Table \ref{tab:lms_vs_cloze}, adding cloze surprisal to a regression already including GRNN, JRNN, Transformer-XL, GPT-2, or BERT surprisal improves their fit. By contrast, human cloze surprisal does not improve regressions already including surprisals from GPT-3, RoBERTa, or ALBERT.

In sum, surprisal calculated using GPT-3, RoBERTa, or ALBERT provides a better fit to N400 data than human cloze surprisals based on analyses in both directions, and BERT surprisal explains some variance in N400 amplitude not explained by human cloze surprisals.

\section{General Discussion}
In this study, we investigated whether linguistic predictions from language models or from human participants better predict the amplitude of the N400, a neural index of processing difficulty. We find that, across the board, the surprisal of three transformer LMs, GPT-3, RoBERTa, and ALBERT, are better predictors of N400 amplitude than cloze. This is consistent with prior work showing the correlation between LM surprisal and N400 amplitude \cite{frank_2015_ERPResponseAmount,aurnhammer_2019_EvaluatingInformationtheoreticMeasures,michaelov_2020_HowWellDoes,michaelov_2021_DifferentKindsCognitive,merkx_2021_HumanSentenceProcessing}. However, to the best of our knowledge, the present study provides the most convincing evidence to date that LM surprisal can outperform cloze as a predictor of N400 amplitude. 

In contrast to a recent large-scale experiment and meta-analysis by Brothers and Kuperberg \cite{brothers_2021_WordPredictabilityEffects}, our results do not show that raw cloze probability is a better predictor of language processing difficulty amplitude than cloze surprisal---in fact, if anything, cloze surprisal is the better predictor. Whether this is because there is a difference in how the N400 and the behavioral metrics analyzed by Brothers and Kuperberg \cite{brothers_2021_WordPredictabilityEffects} relate to word predictability or because of some other difference between the studies is a question for further research.

The skeptical reader might question whether there was some feature of our stimuli that offers an unfair advantage to the LMs over cloze measures. We find this unlikely, given that we have endeavoured to provide a `level playing field'. First, unlike previous work that showed LM surprisal values provide a good account of N400 elicited by different kinds of semantic stimuli equated for cloze probability \cite{michaelov_2021_DifferentKindsCognitive}, the present study involved the experimental manipulation of the predictability of the words. There were no experimental conditions that were matched for cloze but that differed in some other systematic way. Thus, N400 amplitude variance in this study is almost exclusively due to differences in predictability. Second, all zero-cloze items were removed, meaning that any variation between items in terms of predictability was captured by both cloze and LM surprisal. Finally, we included both cloze probability and cloze surprisal as possible predictors to account for the possibility that one might be a better predictor than the other. In summary, the conditions of this study were maximally favorable towards cloze; and yet we see that even so, distributional information can better predict N400 amplitude.

\subsection{Theoretical implications}

Our main result is that overall, GPT-3 surprisal, RoBERTa surprisal, and ALBERT surprisal were each found to be better predictors of N400 amplitude than cloze surprisal values gathered from human participants. While it is striking that cloze probability and surprisal values from a mere 30 participants provide a better fit to N400 data than do surprisal values from GRNN, JRNN, Transformer-XL, and GPT-2, we find that they do not explain any variance in N400 amplitude above and beyond that explained by GPT-3, RoBERTa, and ALBERT surprisal. Furthermore, the surprisal of these LMs, as well as BERT, explain variance in N400 amplitude not captured by cloze. When comparing LMs of the same type, our results also provide new evidence that supports the idea that LMs of higher quality perform better at modeling the N400 and other measures of online human sentence processing difficulty \cite{frank_2015_ERPResponseAmount,goodkind_2018_PredictivePowerWord,aurnhammer_2019_EvaluatingInformationtheoreticMeasures,merkx_2021_HumanSentenceProcessing}. When compared by perplexity, a common evaluation metric for autoregressive transformer LMs, GPT-3 outperforms Transformer-XL and GPT-2 \cite{dai_2019_TransformerXLAttentiveLanguage,radford_2019_LanguageModelsAre,brown_2020_LanguageModelsAre}. Similarly, ALBERT and RoBERTa each out-perform BERT at the GLUE benchmark \cite{wang_2019_GLUEMultiTaskBenchmark}, which covers a wide range of natural language understanding tasks. Finally, all transformer LMs included in this analysis outperform the RNNs (GRNN and JRNN), replicating previous work that transformer LMs are better predictors of N400 amplitude than RNNs \cite{merkx_2021_HumanSentenceProcessing,michaelov_2021_DifferentKindsCognitive}. 

This finding may offer additional insight into why our results diverge from previous behavioral studies showing that cloze probability \cite{brothers_2021_WordPredictabilityEffects} and cloze surprisal \cite{smith_2011_ClozeNoCigar} are better predictors of processing difficulty than LM surprisal beyond the fact that the N400 and behavioral metrics of processing difficulty are not necessarily always comparable. The most sophisticated LM used in these studies is the JRNN (in \cite{brothers_2021_WordPredictabilityEffects}), with n-grams also being used \cite{smith_2011_ClozeNoCigar,brothers_2021_WordPredictabilityEffects}. Thus, our results are actually in line with such findings---in the present study, cloze probability and surprisal out-perform JRNN surprisal at predicting N400 amplitude. Our key finding is that more sophisticated, higher-quality LMs out-perform cloze---as LMs continue to advance and improve, their predictions appear to more closely match those of humans. Thus, our current best operationalizations of predictability based on the statistics of language are the best operationalizations of the predictions underlying the N400 response, and based on the present study, they may continue to get closer.

Until the present study, cloze has been the gold-standard method of operationalizing predictability, and, when tested, the best correlate of behavioural predictability effects \cite{smith_2011_ClozeNoCigar,brothers_2021_WordPredictabilityEffects}. Thus, because the N400 is sensitive to manipulations that cannot be operationalized by cloze probability, it has been argued that it may be more productive to think of the N400 as reflecting `preactivation' \cite{kuperberg_2020_TaleTwoPositivities}, or the `neural activation state at the time a physical stimulus is encountered' \cite{delong_2020_ComprehendingSurprisingSentences} rather than prediction \textit{per se}. For example, besides its high degree of sensitivity to cloze probability, the amplitude of the N400 is also sensitive to factors ostensibly related to the organization of semantic memory. Consider the following set of stimuli from Ito et al. \cite{ito_2016_PredictingFormMeaning}:

\begin{quotation}
\noindent Jack studied medicine in a university and works as a \textbf{doctor/patient/tenant} now. 
\end{quotation}

Here, \textit{doctor} is the highest-cloze continuation of the sentence, while both \textit{patient} and \textit{tenant} have a cloze probability of zero. However, despite the fact that \textit{patient} and \textit{tenant} are equally unpredictable and equally implausible continuations of the sentence (as judged by participants in their study), \textit{patient} elicits a smaller (less negative) N400 than \textit{tenant}. This is one example of a range of studies where words that are semantically related to the preceding context (i.e. \textit{medicine}) or to the most expected continuation of a sentence (i.e. \textit{doctor}) elicit smaller N400 responses than semantically unrelated words, even when matched for cloze \cite{ito_2016_PredictingFormMeaning,thornhill_2012_LexicalConceptualAnticipation,metusalem_2012_GeneralizedEventKnowledge}. Based on such experiments, it has been proposed that implausible continuations like \textit{patient} are `collaterally facilitated' by the preceding context \cite{delong_2020_ComprehendingSurprisingSentences}, or, alternatively, that their preactivation is caused by a separate associative system \cite{frank_2017_WordPredictabilitySemantic}.

However, recent work shows that the difference in N400 amplitude reported in Ito et al.'s \cite{ito_2016_PredictingFormMeaning} study can be successfully predicted based on GRNN and JRNN surprisal \cite{michaelov_2020_HowWellDoes}. This suggests that manipulations thought to be separate or dissociable from predictability---in this case, semantic relatedness to the highest-cloze continuation---may be reducible to an appropriate measure of predictability. That is, \textit{patient} and \textit{tenant} are not in fact equally predictable, and the belief that they are is an artifact of cloze task. If even the GRNN and JRNN, which are among the worst-performing models in the present study, are able to successfully differentiate the predictability of \textit{patient} and \textit{tenant} \cite{michaelov_2020_HowWellDoes} without semantics learned explicitly or through experience of the world, this suggests that humans may not need to rely on such information for prediction either, at least within the N400 window.

The results of the present study may help to illuminate the functional significance of the N400 component by providing evidence for a unified explanation for its sensitivity to what seem to be disparate sources of contextual information. In previous work, we see that semantic relatedness, previously thought to be dissociable from predictability, can successfully be operationalized with LM surprisal \cite{michaelov_2020_HowWellDoes,michaelov_2021_DifferentKindsCognitive}. In the present study, we see that predictability, previously thought to be best operationalized with cloze probability, can be operationalized with LM surprisal, with the highest-quality LMs providing a better operationalization than cloze probability or cloze surprisal. Together, these results suggest that there may be something about the surprisal of high-quality LMs that makes them so well-suited to capturing the predictions of the neurocognitive system underlying the N400 response. LMs are systems trained to predict a word given its context based on the statistics of language. Their degree of success at predicting N400 amplitude relative to other approaches suggests that we should seriously consider that as part of language comprehension, humans may be doing the same.

\subsection{Methodological implications}

Our finding of the relationship between N400 amplitude and surprisal values from GPT-3, RoBERTa, and ALBERT has clear methodological implications. In future work, it may be advantageous for ERP language researchers who want to measure or control the predictability of their stimuli to use surprisal values from these LMs in addition to, or even instead of, cloze probability. As previously discussed, using cloze probability has several theoretical issues, but there are also practical reasons for favoring LM surprisal. For example, it is is easy to gather surprisal values for large stimulus sets (e.g. for every word in a collection of multiple sentences), while this may not feasible for cloze. Additionally, the precision of cloze probability is limited by the number of participants used for the cloze task---with a limited number of participants, small differences in predictability may not be reflected in cloze, and further, this means that even with a large number of participants, variation in the predictability of zero-cloze items may not be detected. LM surprisal, by contrast, allows the researcher to differentiate between items even with a very low probability, making it possible to control for predictability over a wider range than does cloze probability.

However, in addition to these already-known reasons for preferring LM surprisal to cloze, the results of the present study provide another, stronger argument for using LM surprisal over cloze. Even for stimuli that vary in measurable ways in terms of cloze, the surprisals calculated from GPT-3, RoBERTa, and ALBERT's predictions provide a better fit to the N400 data, suggesting that they may better operationalize the predictability underlying the variance in the N400 response to stimuli. Indeed, as discussed, given that these are the highest-quality models tested, we might expect that LM surprisal's ability to capture predictability may continue to improve. ERP language researchers already use other measures derived from linguistic corpora to control their language materials. For example, since the report that corpus-derived metrics of word similarity are correlated with N400 amplitude \cite{chwilla_2005_AccessingWorldKnowledge,parviz_2011_UsingLanguageModels,vanpetten_2014_ExaminingN400Semantic,ettinger_2016_ModelingN400Amplitude}, many researchers have constructed their stimuli such that they are either matched in terms of these metrics, or include similarity metrics as covariates in their statistical analyses \cite{chwilla_2007_ImmediateIntegrationNovel,kuperberg_2020_TaleTwoPositivities,nieuwland_2020_DissociableEffectsPrediction}. The present study suggests that surprisals derived from high-quality LMs should be used analogously in ERP investigations of language processing.

\section{Conclusion}

Previous work has shown that LM predictions correlate with N400 amplitude when cloze does not \cite{michaelov_2020_HowWellDoes,michaelov_2021_DifferentKindsCognitive}. The present study has shown that even in conditions maximally preferable for cloze, LM predictions correlate better with N400 amplitude. Thus, at least in terms of relative strength, the kinds of predictions made by LMs resemble the kinds of predictions made by humans as part of online language comprehension. Thus, the language comprehension system, or at least, the neurocognitive system underlying the N400 response, appears to be more finely attuned to the regularities in the statistics of language than previously thought.


%


\section*{Acknowledgment}
The authors would like to thank Mante Nieuwland and collaborators for making their stimuli and data available online. The authors would also like to thank the anonymous reviewers for their helpful comments.

\ifCLASSOPTIONcaptionsoff
 \newpage
\fi



\bibliographystyle{IEEEtran}
\bibliography{bibliography}
%



%




\end{document}